\documentclass[sigconf,screen]{acmart}

\usepackage{times}
\usepackage{amsmath}
\usepackage[utf8]{inputenc}   
\usepackage[T1]{fontenc}
\usepackage{textcomp}
\usepackage{url}
\usepackage{array} 
\usepackage{pifont}

\usepackage{tabularx}
\usepackage{enumitem}

\usepackage{tabularx}
\usepackage{booktabs}

\usepackage[english]{babel}
\usepackage[autostyle, english = american]{csquotes}
\MakeOuterQuote{"}

\AtBeginDocument{%
  }

\setcopyright{acmlicensed}
\copyrightyear{2026}
\acmYear{2026}
\acmDOI{XXXXXXX.XXXXXXX}
\acmConference[ICAIL]{International Conference on AI and Law}{June 12, 2026}{Singapore}
\acmISBN{978-1-4503-XXXX-X/2018/06}

\begin{document}

\title{On Wednesdays, We Ask Questions: Optimizing "Active Listening" in Automated Legal Triage and Referral}


\author{Quinten Steenhuis}
\affiliation{%
  \institution{Suffolk University Law School}
  \city{Boston}
  \country{United States of America}
}
\email{qsteenhuis@suffolk.edu}

\author{Jacqueline Harvey}
\affiliation{%
  \institution{Suffolk University Law School}
  \city{Boston}
  \country{United States of America }
}

\renewcommand{\shortauthors}{Author 1443}

\begin{abstract}
  The FETCH classifier generates follow-up questions to help refine the best match for the applicant's legal problem, using a low-cost ensemble of LLMs. In this paper, we describe an expert attorney and LLM-assisted evaluation of the follow-up question approach in FETCH and show that while low-cost LLMs perform well at classification tasks, generating high-quality plain-language questions in this setting appears to require a more sophisticated and higher-cost model. Through discussion with legal intake workers, we propose a rubric for the evaluation of legal intake classification questions, and we find that prompt engineering alone is not enough to improve question quality for intake purposes. We also find that LLM-as-judge and human ratings diverge. We demonstrate that with the addition of a single high-cost model, GPT-5, the classifier can elicit relevant information from applicants for legal help, and that the questions lead to more accurate performance at classification tasks. We also find uneven fact elicitation across different categories, including domestic violence, at odds with family law screening protocols, suggesting the value of including dedicated screening panels for certain areas of law.
\end{abstract}




\maketitle
\section{Introduction}
As a new attorney, one of our authors' first legal intakes was with someone who swore that she was being harmed by a voodoo doll. Our author, in their first weeks as an attorney, was surprised and unsure what to do, but pressed on with the intake. While the first description might have seemed like either a joke or something outside of the law's protection, after our author asked enough questions, they learned that the new client was actually dealing with violent threats by a housemate. She lacked the legal words, but it was a legal problem. Once the right framing was in place to understand the client's needs, our author helped the woman get a restraining order.

Legal intake is particularly challenging for self-represented litigants, who make up about two-thirds of civil court litigants in the United States\cite{NCSCCampbell2025Trends}.\footnote{No international data is available, although data from \cite{world_justice_project_global_2019} shows that a similar percentage, about two thirds of the world population, experience significant access to justice problems.} These users often struggle to identify and articulate legally relevant facts, which can lead to inefficiencies in case processing and make it more difficult for legal professionals to assess claims at the earliest stages \cite{ajmi_modernizing_2025}. This suggests that intake systems cannot rely on a single, upfront description of a user's problem because key legal details may only emerge through follow-up questioning. Iterative classification, where follow-up questions help surface relevant details that may not appear in the initial narrative, is an important part of human intake.

Intake systems often force applicants to articulate their legal issue at the beginning of an intake, even though applicants may not know which details are necessary to share, may include information that is irrelevant, or may leave out key facts.  The FETCH\cite{steenhuis2026fetch} classifier is an ensemble classifier that helps match an applicant's natural language description of their problem to a legal taxonomy, and then to match the applicant with an attorney. The FETCH classifier also uses an ensemble approach to generate and then merge questions to help improve the classification of the applicant's legal problem. This paper focuses on:

\begin{itemize}
    \item Factors that impact the quality of generated questions.
    \item Whether generated questions can improve classification accuracy.
\end{itemize}

During the course of this study, we held a small focus group with legal intake staff at the Oregon State Bar, evaluating the quality of the generated questions and other aspects of the legal referral tool. From this, we created a rubric and designed an experiment to determine if refinements to the prompt could improve the quality of the generated questions without structural revisions to the tool.

Like the woman in our author's first intake, applicants for legal help to a general referral platform are often in vulnerable situations, facing urgent legal problems such as eviction, domestic violence, or loss of income. In this stressful context, the use of plain language is important to ensure that applicants understand the questions being asked of them \cite{steenhuis_beyond_2023, welsh_effects_2013}. This emerged as a key area of interest in the focus group.

Another concern that emerged in the focus group was relevance. Broadly speaking, relevance meant generating questions that helped the applicant narrow their problem down to the appropriate classification without asking for unnecessary information.

With the focus group feedback in mind, we evaluate the quality of the generated questions with an LLM-graded rubric as well as expert legal human evaluators across the two dimensions of relevance and readability. We tested question generation with the FETCH\cite{steenhuis2026fetch} dataset, together with a small set of novel human-generated questions created by legal intake staff drawn on past experience and adapted from live user queries.

Finally, we explore changes in the quality of the generated questions through small changes in the prompting strategy, including the addition of few-shot examples into the prompts, and explore whether using a more capable LLM can address the problems that cannot be solved by prompt engineering alone. Finally, we provide preliminary evidence that high-quality questions can improve the quality of classification.

\subsection{Research questions}
\begin{itemize}
    \item RQ1: Can LLMs be used in an ensemble approach to generate high-quality clarifying questions for legal issue classification tasks?
    \item RQ2: Can LLM model-graded evaluations assist with improving the generation of clarifying questions?
    \item RQ3: Do clarifying questions improve the performance of LLM-assisted classification in the legal context?
\end{itemize}
\section{Prior work}
\subsection{AI for access to justice}
AI, including rules-based expert systems and large language models, increasingly has a central place in solving access to justice problems with technology. See, e.g., applications to legal information tools \cite{branting2001advisory, westermann_justicebot_2023}, form-filling and expert systems \cite{steenhuis_digital_2021,  steenhuis2024ai, westermann2024dallma}, and dispute resolution \cite{ Westermann_thesis_2023, bickel2015online, Branting2022-BRAACM-5} as cited in \cite{steenhuis2026fetch}.
\subsection{AI for generating questions and qualifying leads}
Automatic question generation is a core functionality of modern LLMs \cite{Mulla2023}, but is a field with a long history, with prior work discussing rules-based approaches and retrieval using semantic similarity. Other approaches, such as logistic regression and random forest,
have been applied to the broader task of qualifying sales leads \cite{nair2020enhancing}. Pre-written questions have also been applied to legal intake \cite{remus2017robotlawyer}.
\subsection{Measuring form usability for access to justice}
Steenhuis, Willey and Colarusso \cite{steenhuis_beyond_2023} proposes five factors that impact the usability of a form for a self-represented litigant: the readability of the text, the ability to provide an accurate response, the ability to provide a complete response, and the burdens, both time and psychological trauma related, that providing a response imposes on the litigant. Jarrett and Gaffney \cite{jarrett_forms_2008} further explore specific input choices, such as the use of radio buttons and checkboxes over open-ended questions, that can reduce the user's burden when providing a response and increase the response's accuracy.

GDPR requires that intake questions are "adequate, relevant, and limited" \cite{gdpr_art5_processing_personal_data}, consistent with recommendations from the Nielsen Norman Group on survey design \cite{brown_2023_survey_best_practices} (suggesting that you only ask questions that you need answered) and the American Bar Association's guide on legal intake \cite{ABA2026ClientIntake}, recommending avoiding "excessive detail".

\subsection{AI for measuring and improving readability of legal text}
Early research in measuring readability, including \cite{Flesch_1948_NewReadabilityYardstick, ChallDale1995}, focused in large part on easily measured mechanical attributes of the text, such as sentence length and the number of syllables in each word, or those attributes combined with vocabulary in the case of \cite{ChallDale1995}. These metrics have been criticized as overly simplistic and inadequate to capture actual readability \cite{redish_readability_2000, Benjamin2012ReconstructingReadability}. Machine learning approaches such as \cite{sepehri2023passivepy, francois_amesure_2020}, and LLMs \cite{Novotna2026PONK} (exploring LLMs' ability to measure lexical surprisal) have expanded the flexibility of measuring readability beyond these limited mechanical features.

Our work focuses on generating follow-up questions for legal intake classification, which involves aspects of both practical questions of readability, with an emphasis on readability, ease of providing a response, and the relevance of the question. In the context of the FETCH classifier specifically, only questions that help match the applicant to the right source of legal help are relevant, while in other forms of legal intake, the goal might be to obtain a complete picture of the applicant's legal situation to help build a case or a defense.

\section{Background}
Intake is a key phase in legal representation,\footnote{For example, the 2023 Clio Legal Trends Report includes a survey of lawyers across North America, showing that lawyers spend up to 33\% of their time on business development \cite{ClioLegalTrends2023}, including lead qualification.} and may be considered a subset of the broader task of "qualifying" a lead \cite{nair2020enhancing}. At the intake stage of a case, the applicant shares information with a legal office about their case. The attorney or paralegal has to play an active listening role. The questions they ask must direct the applicant to providing relevant information about their experience of a problem. They match questions to specific details of legal claims and remedies, while being sure to avoid asking the applicant to repeat themselves or to provide detail that isn't important to their case.

The applicant often does not know which part of their story is legally relevant. Self-represented litigants, who comprise a substantial majority of users in many legal systems, frequently struggle to identify and articulate key legal facts, leading to procedural inefficiencies and delays in case processing \cite{ajmi_modernizing_2025}. In response, structured intake systems increasingly relied on guided, question-driven triage processes to support early issue identification and route cases to appropriate legal pathways. These systems use iterative questioning to assess factors such as case complexity, urgency, and risk, recognizing that initial user narratives are often incomplete or misaligned with legal categories \cite{ajmi_modernizing_2025}.  However, empirical evaluations of such systems highlight significant challenges in their implementation. Usability studies of court-based triage tools show that users frequently struggle to understand legal terminology, interpret question intent, and navigate question flow, often leading to confusion,  hesitation, or disengagement during the intake process \cite{pratt_johnson_2024}. Users also report uncertainty about how their responses will be used and difficulty understanding the purpose of individual questions, particularly when those questions address sensitive topics \cite{pratt_johnson_2024}. These findings also suggest that question quality must be evaluated not only in terms of correctness, but in terms of how users understand and respond to questions in context. In practice, even technically accurate questions may fail if they do not match user expectations or follow a clear conversational progression. This highlights the importance of designing follow-up questions that reflect an active listening approach, rather than simply extracting additional information.

In this context, the intake worker's questions may be key to deciding whether a user has a viable legal claim or even a legal issue at all. In other cases, they may have a solution but the attorney who they are speaking with may not be able to help with their problem.

Automatic question generation has the potential to play the active listening role that rules-based systems fail at. By generating clarifying questions, such systems can reduce the time and effort required to design intake flows across diverse legal domains. On the other hand, irrelevant, repetitious, or confusing questions may frustrate, annoy,
or cause the applicant to give up before getting a legal referral.

\section{Data}
We use a de-duplicated subset of the FETCH dataset\cite{steenhuis2026fetch}, with an additional 60 questions captured through experimental, ad hoc testing by staff at the Oregon State Bar. These questions used modified examples drawn from their referral experience as well as exploring edge cases. For example: "An alien stole my cat and wife" and "My neighbor won't stop playing his clarinet all night." We also removed queries from this dataset that were not written in the English language.

\subsection{Focus group and qualitative feedback results}
In three Zoom breakout rooms, we spent an hour and a half with 7 employees of the Oregon State Bar to gather qualitative feedback about an early version of the referral tool. Applicants were invited to test queries of their own invention.

Outside of the focus group, staff conducted additional ad hoc testing of the tool, again,
employing a mix of questions drawn from recent referrals and edge cases that occurred to
the staff to test. From this ad hoc testing, we expanded the test question dataset and 
took note of the additional feedback from a U.S. State Bar's staff.

User feedback focused on a few dimensions:
\begin{itemize}
    \item Tricky legal vocabulary and acronyms, such as EEOC (Equal Employment Opportunity Commission), DUII (Driving Under the Influence of Intoxicants), or SSDI (Social Security Disability Insurance), which sometimes came from the taxonomy keywords.
    \item Duplicative questions, either shown at the same time or on a follow-up screen.
    \item Questions about whether the person wants a lawyer, or specific details about the kind of help they want, which were not necessarily helpful at choosing between two different legal categories.
    \item Questions that didn't demonstrate "active listening"--that is, that re-asked information that seemed to clearly be contained in the original query.
\end{itemize}

Some feedback would be difficult to address in the constraints of the tool. For example, one query asked "My mom is letting her ex-husband live with her and I want him to leave the house." Staff were concerned that the user would not have standing to sue or remove the ex-husband from the home, but screening out an applicant in this situation with an automated determination may present ethical concerns. Similarly, focus group participants explored situations that would be triaged or escalated to emergency services in in-person referral, such as an applicant using threatening language. We decided to note this feedback but did not act on it in this experiment.\footnote{We noted that OpenAI's moderation endpoint might be useful to help filter and triage dangerous applicant intent, but also that it would take careful evaluation to determine if the accuracy of this was high enough to avoid screening out people reporting on dangerous situations but with positive intent. For example: we were concerned that flagging violence would screen out people reporting abuse or violence that they were the survivor of. Even this constraint is problematic: applicants who were the aggressors in a past incident are still entitled to hire an attorney, making this a very difficult guardrail to implement.}

\section{Methodology}
\begin{figure}
    \centering
    \includegraphics[alt={Diagram of FETCH question generation, showing progression from user problem description to final set of questions.},width=0.75\linewidth]{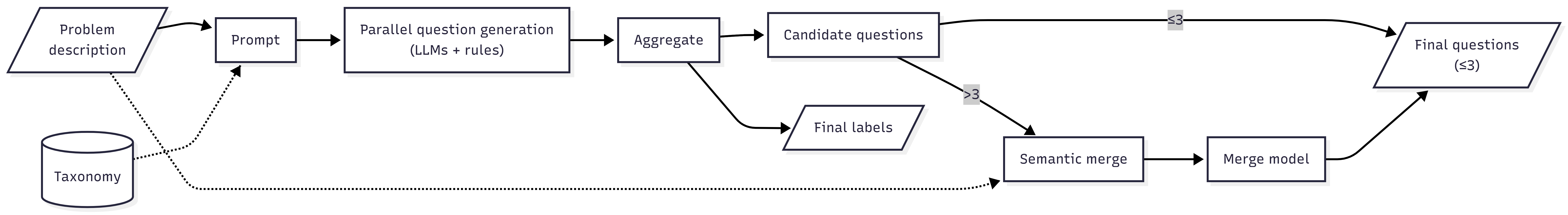}
    \caption{Diagram of FETCH question generation, showing progression from user problem description to final set of questions.}
    \label{fig:flowchart_diagram}
\end{figure}

As described in \cite{steenhuis2026fetch} and Figure \ref{fig:flowchart_diagram}, FETCH uses an ensemble of three low-cost LLM providers to generate candidate follow-up questions, which are then semantically merged and adjusted to meet quality standards using a fourth LLM. In this experiment, we modified both the shared question generation prompt and the merging prompt to improve the quality of the generated questions. We also experimented with a "stronger" LLM (one with more parameters) to see if this had a substantial impact on question quality.

\begin{itemize}
    \item GPT-5-nano, gemini-2.5-flash-lite, and mistral-small-7b were used to generate candidate questions in experiments one and two, while GPT-5 was added in experiment three. Temperature is set to 0 for gemini and mistral models. The GPT-5 family, which does not have a temperature setting, had its thinking level set to "low".
    \item GPT-5-nano was used for semantic merging in experiments one and two, while GPT-5 was used in experiment three. Thinking level was set to "low" for semantic merging as well.
\end{itemize}

\subsection{Experimental setup}
We evaluated the changes over three experiments, incorporating the feedback obtained during
the focus group and ad hoc testing sessions. Our baseline condition reflects the original prompts as described in \cite{steenhuis2026fetch}, with minor revisions to the semantic merging prompt that were made prior to the start of this study. 

Promptfoo setup and raw results are available at \url{https://github.com/nonprofittechy/followup-study-paper-repo}.

\subsubsection{Experiment 1: Prioritizing readability}
The prompt was changed to emphasize readability as the first listed goal in semantic merging. We put ``Simplify the language to improve plain language readability'' as an instruction in the semantic merge prompt before ``Merge similar questions'' and ``Remove questions \ldots''. The  change also added: ``Add explanations of legal jargon when helpful. Use simpler vocabulary terms to make your phrasing colloquial. Use simple syntax, including active voice. Avoid nominalizations and other complex phrasing.'' In parallel, the question generation prompt was tightened from ``Use plain language and target a 6th grade reading level'' to ``Use plain language, including simple vocabulary terms like `get' instead of obtain, `want' instead of `seeking', etc.''

\subsubsection{Experiment 2: Explicit examples to clarify abstract rules}
Building on Experiment 1, Experiment 2 added three explicit, verifiable constraints and few-shot feedback:

\begin{enumerate}
    \item Glossing: ``When you must use a legal term, ALWAYS add a plain-English explanation in parentheses: Examples: `guardianship (legal right to make decisions for someone)', `executor (person who handles the estate)', `foreclosure (losing your home because of unpaid mortgage)'.''
    \item Substitution rules: ``Use these simpler words instead of formal legal terms: \ldots `found guilty' instead of `convicted' \ldots `court order to stay away' instead of `restraining order' \ldots''
    \item Redundancy detection: ``Before including a question, check: did the user already answer this? If yes, do NOT ask it. Example: If user says `my landlord won't return my deposit', do NOT ask `Is this about a security deposit?'\,'' Finally, the prompt added explicit bad-question examples with failure analysis: ``[BAD] `Was the defendant convicted of a felony?' (Jargon: `defendant', `convicted', `felony' — user may not know these)''.
\end{enumerate}

\subsubsection{Experiment 3: measuring effect of improved questions on model classification}

In order to simulate a human interacting with the FETCH classifier, we created a new synthetic dataset based on the Oregon State Bar's provided sample user problem scenarios. In this synthetic dataset, we broke the original scenarios into two components: a setup and a hidden fact. With the help of Claude Sonnet 4.6 (a model not used in classification), we then generated a set of 200 tricky "flip" scenarios that we predicted should lead to different or opposite classification according to the OSB taxonomy, based on variations in the key facts. An experienced attorney reviewed the synthetic scenarios for accuracy.

Because of the disappointing results from prompt engineering alone, during the course of this project, we swapped the GPT-5-nano model in both question generation and semantic merging for GPT-5. That remained the baseline condition in experiment 3. The GPT-5 family includes GPT-5, GPT-5-mini, and GPT-5-nano, with the models decreasing in both performance on benchmarks and cost in the same order \cite{openai2025gpt5developers}. Thinking remained set to "low".

Table \ref{tab:classification-flip-examples} shows a small sample of the "flip" scenarios:
\begin{table*}[t]
\centering
\scriptsize
\setlength{\tabcolsep}{4pt}
\renewcommand{\arraystretch}{1.15}
\begin{tabularx}{\textwidth}{@{}p{0.11\textwidth} X p{0.21\textwidth} p{0.21\textwidth} X@{}}
\toprule
\textbf{Scenario ID} &
\textbf{Opening query} &
\textbf{Initial classification} &
\textbf{Updated classification} &
\textbf{Hidden fact causing flip} \\
\midrule
p05\_AtoB\_04 &
I told someone we wouldn't be needing their help anymore, and now they're demanding payment for work they haven't done yet. &
Labor \& Employment: Wrongful Discharge -- Employee &
Business and Corporate: General (contracts, entities) &
I run a catering business and hire chefs on a per-event contract basis. \\

p09\_BtoA\_09 &
I have a judgment from small claims court but still haven't been paid. What's the next step? &
Debtor/Creditor: Judgment Collection &
Bankruptcy: General (Bankruptcy) &
I was told I can't collect because I'm going through bankruptcy myself. \\

p01\_BtoA\_06 &
I hurt my knee and I’m in a lot of pain lately. I think I might have torn something. &
General Litigation: Personal Injury &
Workers' Comp: General (Workers' Comp) &
The injury happened while I was stocking shelves at my part-time job. \\

p02\_AtoB\_09 &
I need advice on how to separate from my partner. We’ve had a lot of disagreements lately. &
Family Law: General (Divorce/Separation etc.) &
Family Law: Domestic Violence &
My partner cut off my access to our bank account and won’t let me have money. \\

p06\_BtoA\_03 &
I’m having trouble getting paid back for a service I did for someone. &
Debtor/Creditor: General (Creditor) &
Consumer Law: General Consumer &
The person said I didn’t finish the job properly, so they aren’t paying. \\
\bottomrule
\end{tabularx}
\caption{Randomly sampled classification flip scenarios.}
\label{tab:classification-flip-examples}
\end{table*}

We then used FETCH in a two-step classification pipeline with Promptfoo as follows:

\begin{enumerate}
    \item The initial truncated scenario was provided to FETCH for classification.
    \item If FETCH generated questions, GPT 4.1 was used to determine if the hidden fact was a relevant response to the generated question, with temperature set to 0.
    \item If the fact was determined to be relevant by GPT 4.1, the combination of truncated scenario, generated question, and hidden fact response were provided to FETCH for a second classification attempt.
    \item Finally, Promptfoo compared the final classification to the predicted classification.
\end{enumerate}

\subsection{Evaluation rubric}
Our rubric encompasses the two dimensions of relevance and readability. These dimensions are consistent with the general rubric for legal AI help proposed in \cite{hagan2023good} as well as the factors defined in \cite{steenhuis_beyond_2023, jarrett_forms_2008, gdpr_art5_processing_personal_data, brown_2023_survey_best_practices, ABA2026ClientIntake}.

We say that a question is \textbf{relevant} if:
\begin{itemize}
    \item The question, if answered, would add information that would lead to a more accurate classification of the user's legal problem.
    \item It does not ask the user to repeat information that the user has already provided.
    \item It does not ask about something unnecessary to the classification, such as which venue or kind of relief the user is looking for.
\end{itemize}
We say that a follow-up question is \textbf{readable} if:
\begin{itemize}
    \item It uses common vocabulary words, and it does not introduce an uncommon vocabulary word unless the user already used it.
    \item It does not include passive voice or other complex grammatical syntax.

\end{itemize}

We asked both an expert attorney who is familiar with legal intake and an LLM to score the 416 example applicant queries in our dataset using the standards above.

\subsection{Model-graded evaluation}
We used Promptfoo as a model-graded evaluator in experiments 1 and 2, with gpt-5-nano as the LLM-as-judge.

\section{Results}
Prompt iteration substantially improved performance under the evaluation of the LLM-as-judge as shown in Table~\ref{tab:summary-results}.
\begin{table}[t]
\centering
\caption{Summary of LLM-Graded Follow-up Question Evaluation Results}
\label{tab:summary-results}
\begin{tabular}{lccc}
\toprule
\textbf{Run} &
\textbf{Passing} &
\textbf{Total} &
\textbf{Pass Rate} \\
Baseline & 346 & 416 & 83.17\% \\
Experiment~1 & 349 & 416 & 83.89\% \\
Experiment~2 & 374 & 416 & 89.90\% \\
\bottomrule
\end{tabular}
\end{table}

Overall, with the changes in Experiment 2, 89.9\% of the generated question sets met the criteria, as defined by the model-graded judge with our two-part rubric. As discussed below, when considering only relevance, the passing score raises to \textbf{97\%}.

\subsection{Human binary comparison of results in experiments 1 and 2}
We asked an experienced attorney with past work in intake to perform a binary preference evaluation of 146 of the questions generated in experiments 1 and 2 (prompt engineering without changing model strength), with mixed results shown in Table \ref{tab:expert_preference}. The rater was asked to consider the same relevance and readability factors as given to the LLM-as-judge when selecting between the two results.

\begin{table}[h]
\centering
\begin{tabular}{l r}
\hline
\textbf{Expert preference} & \textbf{Count} \\
\hline
Experiment 1 question set & 79 \\
Experiment 2 question set & 64 \\
Neutral / unable to decide & 3 \\
\hline
\textbf{Total} & \textbf{146} \\
\hline
\end{tabular}
\caption{Expert overall preference between question phrasing in Experiment 1 and Experiment 2.}
\label{tab:expert_preference}
\end{table}

\subsection{Human evaluation of experiment 3}
Because this experiment involves active work with the Oregon State Bar, we asked for their qualitative feedback on the results of experiment 3. In conversations, the team reported being very happy with the improved readability, relevance, and accuracy on tougher classification edge tasks after the model switch to GPT-5. We plan to return to these data to perform a more detailed quantitative analysis that directly compares question readability.

\subsection{Evaluation of the effect of follow-up questions on classification accuracy}

Our classification flip experiment showed that when the litigant had a relevant fact that changed classification, FETCH asked a question that elicited it 69\% of the time. This follow-up question rescued an incorrect classification 4.5 times more often than it degraded a correct classification, showing the value of asking follow-up questions. Detailed results are in Table \ref{tab:outcome-matrix}. 

\begin{table*}[t]
\centering
\caption{Outcome Matrix for Matched Scenarios}
\label{tab:outcome-matrix}
\begin{tabular}{llrr}
\hline
\textbf{Initial} & \textbf{After follow-up} & \textbf{Count} & \textbf{\% of matched} \\
\hline
Correct & Correctly flipped to expected final category & 72 & 52.2 \\
Correct & Unchanged; follow-up neutral and still correct & 26 & 18.8 \\
Correct & Changed to wrong category; true degradation & 4 & 2.9 \\
Wrong & Rescued: correctly flipped to expected final & 18 & 13.0 \\
Wrong & Still wrong after follow-up & 18 & 13.0 \\
\hline
\multicolumn{2}{l}{\textbf{Net impact: rescued $-$ truly degraded}} & \textbf{+14} & \\
\hline
\end{tabular}
\par\smallskip
\footnotesize{\textit{Note:} Matched scenarios are those where the generated follow-up question matched the hidden clarifying fact ($n = 138$). Follow-up was $4.5\times$ more likely to rescue a wrong classification than to corrupt a correct one.}
\end{table*}

Table \ref{tab:per-pair-breakdown} shows that the performance of FETCH was not even across all case types in generating relevant follow-up questions. The largest disparity is in questions about domestic violence, where only 10\% of interactions led to a follow-up question that elicited the relevant domestic violence information. While arguably this question may sometimes seem jarring in an otherwise straightforward divorce case, \cite{rossi2023masics_instrument} recommends a protocol for asking screening questions about domestic violence in all family law intake. This rule is not programmed into FETCH. This is further discussed in Table \ref{tab:notable-failure-modes}.

The classification flip experiment also shows other gaps in question protocol that may show overlapping categories: the model did not distinguish between criminal vs restraining order cases or worker's compensation vs personal injury, nor did it distinguish between employment administrative hearing scenarios rather than general labor and employment. While in each of these cases, the applicant may end up with an attorney who can help with their case, it suggests that triggering some additional panels of questions for certain case types might improve performance in any area where the applicant may not realize or be reluctant to disclose important information.

\begin{table*}[t]
\centering
\caption{Per-Pair Breakdown}
\label{tab:per-pair-breakdown}
\begin{tabular}{lrrrrrr}
\hline
\textbf{Swap Pair} & \textbf{$n$} & \textbf{Init.} & \textbf{Init. \%} & \textbf{Matched} & \textbf{Cov. \%} & \textbf{Final \%} \\
\hline
\texttt{custody\_vs\_support} & 20 & 20 & 100.0 & 17 & 85.0 & 100.0 \\
\texttt{domestic\_violence} & 20 & 20 & 100.0 & 2 & 10.0 & 100.0 \\
\texttt{dui\_vs\_dmv} & 20 & 20 & 100.0 & 20 & 100.0 & 90.0 \\
\texttt{tenant\_vs\_landlord} & 20 & 19 & 95.0 & 19 & 95.0 & 94.7 \\
\texttt{debtor\_vs\_creditor} & 20 & 16 & 80.0 & 11 & 55.0 & 72.7 \\
\texttt{employee\_vs\_employer} & 20 & 16 & 80.0 & 18 & 90.0 & 61.1 \\
\texttt{employment\_admin} & 20 & 13 & 65.0 & 4 & 20.0 & 0.0 \\
\texttt{bankruptcy\_vs\_collections} & 20 & 10 & 50.0 & 9 & 45.0 & 88.9 \\
\texttt{injury\_location} & 20 & 20 & 100.0 & 19 & 95.0 & 26.3 \\
\texttt{criminal\_vs\_restraining} & 20 & 0 & 0.0 & 19 & 95.0 & 15.8 \\
\hline
\textbf{Overall} & \textbf{200} & \textbf{154} & \textbf{77.0} & \textbf{138} & \textbf{69.0} & \textbf{65.2} \\
\hline
\end{tabular}
\par\smallskip
\footnotesize{\textit{Note:} Init. is the initial correct count. Cov. \% is follow-up question coverage. Final \% is final classification accuracy among matched scenarios.}
\end{table*}

\begin{table*}[t]
\centering
\caption{Notable Failure Modes}
\label{tab:notable-failure-modes}
\begin{tabular}{lp{0.58\linewidth}r}
\hline
\textbf{Pair / Direction} & \textbf{Failure pattern} & \textbf{Count} \\
\hline
\texttt{criminal\_vs\_restraining} 
& Model consistently returns ``General Litigation $>$ Stalking Orders'' regardless of relationship context; neither Criminal Law nor Family Law labels are produced 
& 19/19 \\

\texttt{employment\_admin} 
& Model ignores administrative hearing signals even when explicitly told a state-agency hearing is scheduled; stays on Labor \& Employment 
& 4/4 \\

\texttt{injury\_location BtoA} 
& Follow-up confirming contractor/off-clock status does not flip classification from Workers' Comp to Personal Injury 
& 14/19 \\

\texttt{domestic\_violence} 
& FETCH rarely generates a question probing for abuse history, with only 10\% coverage, so the flip opportunity is almost never created 
& 18/20 \\
\hline
\end{tabular}
\par\smallskip
\footnotesize{\textit{Note:} Counts reflect the number of affected matched or total scenarios for each failure pattern, as applicable.}
\end{table*}

\section{Discussion}

The results show that prompt engineering, while helping the model meet strict plain language heuristics, failed to improve the questions sufficiently to meet human preferences. Our human rater, summing up the impact of experiments 1 and 2, said that most of the time she did not like either set of questions. This matches the baseline condition reached in the focus group.

However, qualitative feedback from Oregon State Bar staff suggested that changing one of the ensemble models from GPT-5-nano to GPT-5 improved readability, relevance, and classification behavior.

\subsection{plain-language question generation is a hard task}

The detailed changes required in this experiment contrast to the immediate improvement that ensemble techniques made in improving one-shot
classification accuracy in FETCH \cite{steenhuis2026fetch}, suggesting that question generation
is a more complex task that requires more explicit guidance to the model, as well as a more capable LLM.

This finding is consistent with prior work on LLM-based question generation, which shows that the task involves multiple steps, including aligning with the task goal, refining outputs, and evaluating quality \cite{nikolovski_advancing_2025}. Rather than being a single-step task, effective question generation requires clear prompts and constraints. This supports the observation that question generation requires more guidance than classification tasks. This multi-step view also helps explain the improvements observed in Experiment 2. By adding explicit constraints and examples, the prompt more directly guided the model across these stages, particularly in aligning outputs with the task goal and enforcing consistency in refinement. In contrast, less structured prompts may leave steps under-specified leading to more variable question quality. This also suggests that improvements in question generation may depend less on model choice and more on how well the prompt structure captures these immediate steps.

\subsection{Model-graded evaluation shows slight promise to help with iterative prompt engineering}
Model-graded evaluation allowed us to work on a quick feedback loop and make improvements to the prompt without significant regressions. While model-graded evaluation strongly favored Experiment 2, a human expert showed a slight preference for Experiment 1. This divergence may support the model's bias to its own generated responses, and suggests that additional explicit constraints, such as mandatory glossing and substitution, may improve rubric compliance while subtly degrading qualities valued by intake professionals, such as conversational flow or perceived naturalness. Recent work shows that changes to prompt design can affect model accuracy and compliance, and can change the type of outputs the model produces,  which is consistent with the differences observed between our experimental conditions \cite{atreja_whats_2025}. This helps explain the difference between model and human evaluations in our results. Prompts designed to meet a scoring rubric may produce answers that score well, but feel less natural or conversational. In this set-up, the prompt effectively decides what the system optimizes for, which may not fully  match how people judge question quality in practice.  This additionally underscores the importance of hybrid evaluation approaches. For deployed systems, this suggests that model-graded evaluation is best used as a rapid diagnostic and regression-detection tool, rather than as a sole optimization target.

\subsection{Limitations and risks}

\begin{itemize}
    \item Prompt engineering to improve readability has language and culturally specific features. For example: passive voice is an English-language feature. However, improvements related to relevance are likely to be useful in broader geographical regions, as are instructions to provide explanations (glossing).
    \item Iterative prompt improvement has a risk of overfitting. Careful prompt design to avoid overly specific rules in favor of general purpose improvements that are consistent with the literature on question design makes this less of a concern.
    \item Generating questions for applicants for legal help may risk psychological harm to the applicant. For example: questions about traumatic experiences 
          can lead to re-traumatization \cite{steenhuis_beyond_2023, welsh_effects_2013}, but see \cite{weiss_2025_beyond_retraumatization}, suggesting that questions about traumatic events have only slight negative impact on interview participants.
          \item Other risks are mitigated in part by the inclusion of off-ramps for difficult cases and the inclusion of a dedicated phone number on every screen of the live referral tool, including error screens.
          \item The disparate rate of appropriate follow-up questions, especially related to domestic violence, is a departure from current norms in family law intake, where experts recommend asking screening questions about domestic violence \cite{rossi2023masics_instrument, applegate2025screening_violence}. We will consider adding explicit instructions on screening for certain case types, especially where applicants may be reluctant to self-disclose but the information would be relevant to the referred attorney. This will take careful design to be thoughtful and avoid retraumatization.

\end{itemize}

\section{Conclusion}
Taken together, prior work on legal intake, and LLM behavior suggests that good intake systems need both structure and flexibility. They need to guide users with clear questions, but also account for how users actually understand and respond to those questions. Our results support this, showing that prompt design can improve question quality but also introduce tradeoffs that are not fully captured by automated evaluation alone. Relevant, simple to use, question language is still a challenge for current inexpensive LLMs.
Strategies to improve readability and relevance through prompt engineering show limited promise, at least in model application of a strict rubric and when using lower cost LLMs.

Lack of agreement between our sole human evaluator and the LLM-as-judge is an important preliminary finding. We will continue to explore improvements to the LLM-as-judge prompt to better align results with human preferences, as human rating is extremely time-consuming and use of LLM-as-judge has promise to greatly speed up prompt iteration, development, and catching of regressions when adding new models to the ensemble as old models are deprecated.

Finally, the ability of the FETCH classifier to elicit relevant questions and to act on the new information it elicits is an important finding. This suggests that one-shot classification can be improved upon with the help of follow-up question generation by LLMs at a relatively low cost and complexity.\footnote{As noted in \cite{steenhuis2026fetch}, adding GPT-5 as a model is projected to add \$300/year to the Oregon State Bar's costs, a relatively modest increase.}

\begin{acks}
We used LLMs, including Codex and GitHub Copilot, to format configuration files and format \LaTeX  code for tables and figures. ChatGPT was also used to help provide feedback on unclear or confusing language but not to draft text.
\end{acks}

\bibliographystyle{ACM-Reference-Format}
\bibliography{on_wednesdays}


\end{document}